# Programmable magnetic soft robots with controlled locomotion and directional liquid cargo release

Youyi Zhou[1], Zoe Evelyn Gureno[1], Meghna Majumder[1], Yunus Alapan[1,2,3], *Member, IEEE*

***Abstract*— Magnetically programmable soft elastomers enable complex shape morphing and locomotion dynamics in small scale soft robots under external magnetic fields. Benefiting from their programmed deformation and wireless actuation capabilities, magnetic soft robots have emerged as promising platforms for targeted drug delivery, especially in human gastrointestinal tract. However, achieving controlled directional liquid cargo release toward desired tissue interface while preserving the encoded shape morphing and locomotion capabilities remain a significant challenge. Here, we report a new design strategy that employs an optimized magnetization profile to enable controlled directional release of aqueous cargo without compromising shape morphing and locomotion capabilities. Magnetic soft robots with a specific spatially distributed magnetization profile allow directional alignment of the release interface with the orientation of the external magnetic field. This orientation control ensures active alignment of the release interface toward the intestinal wall prior to drug release. An interconnected microporous elastomer is embedded within the robot for aqueous cargo storage, while a thin microcrystalline wax layer seals the release opening hole to isolate the stored liquid cargo from external environment during transport. Triggered release is achieved by mechanically rupturing the wax sealing layer under a higher-magnitude external magnetic field. Controlled directional flipping, locomotion, and triggered release are decoupled through external magnetic field's direction and strength. The controlled directional release strategy reported here integrates directional targeted liquid cargo release, shape morphing, and locomotion, which establishes the groundwork for target drug delivery in gastrointestinal tract applications.**

## I. Introduction

Shape-programmable magnetic soft robots have attracted substantial interest in the last decade for minimally invasive drug delivery applications due to their wireless actuation and complex locomotion capabilities under external magnetic fields [1-3]. Magnetic soft robots are generally made of polymeric materials, such as silicone elastomers or hydrogel, and inclusions of ferromagnetic micro particles such as $Nd_2Fe_{14}B$ and $CrO_2$ [1, 2, 4, 5]. Magnetic soft robots then can be encoded with spatially varying magnetization profiles either by under a large magnetic field (~>1 T) in a pre-folded state [1] or by digital programming of local voxels that are heated beyond the Curie temperature of embedded particles [6]. Magnetically programmed soft robots experience spatially distributed forces and torques, aligning the magnetization profiles with the external magnetic field direction, resulting in complex shape morphing. Temporal configuration of external fields further enables locomotion of magnetic soft robots under dynamic deformations.

Taking advantage of these unique properties and behaviors, shape-programmable magnetic soft robots have emerged as promising platforms for various biomedical applications [1, 3, 7], particularly for targeted drug delivery in human body, such as gastrointestinal (GI) tract [5, 8, 9, 10]. Nevertheless, the currently available magnetic soft robotic devices for drug delivery require external chambers and intricate loading and release mechanisms, with trades off in sample size and locomotion capabilities. In addition to external chamber-based designs, microporous magnetic soft materials have been proposed for direct loading of liquid cargoes into the material itself with a wax sealing layer for magnetic field-triggered release, which overcomes the limitations in complex shape morphing during liquid cargo transportation [11]. However, targeted and effective drug absorption in GI tract requires continuous localized delivery toward the tissue interface, such as intestinal wall. Despite previous advances, achieving controlled directional drug release or self-correcting orientation toward the intestine wall interface without sensing remains challenging.

In this paper, we introduce a magnetic microporous soft robot with controlled directional release and self-correcting orientation toward intestine wall interface without sensing (Fig. 1a, b). The proposed microporous magnetic soft robot relies on an optimized magnetization profile with orthogonal and transverse magnetization components, which enables self-correcting orientation by flipping the soft robot to align with external magnetic field direction (Fig. 1b). The introduced design not only enables self-correction of release interface at bottom surfaces but also allows flipping and attraction to any arbitrary surface in 3D, including top surfaces, under the guidance of an external permanent magnet. This design also preserves programmable locomotion and shape morphing capabilities. The liquid cargo storage and controlled release is enabled by an interconnected microporous liquid cargo storage unit and a microcrystalline wax thin film for fluidically encapsulating of the liquid cargo from external environment (Fig. 1a).

[1]Y. Zhou, Zoe Evelyn Gureno, Meghna Majumder, and Y. Alapan are with the Bio-integrated Robotics Lab at Mechanical Engineering Department, University of Wisconsin - Madison, Madison, WI 53706 USA (alapan@wisc.edu).

[2]Y. Alapan is with the Biomedical Engineering Department, University of Wisconsin - Madison, Madison, WI 53706 USA.

[3]Y. Alapan is with the Carbone Cancer Center, University of Wisconsin - Madison, Madison, WI 53706 USA.

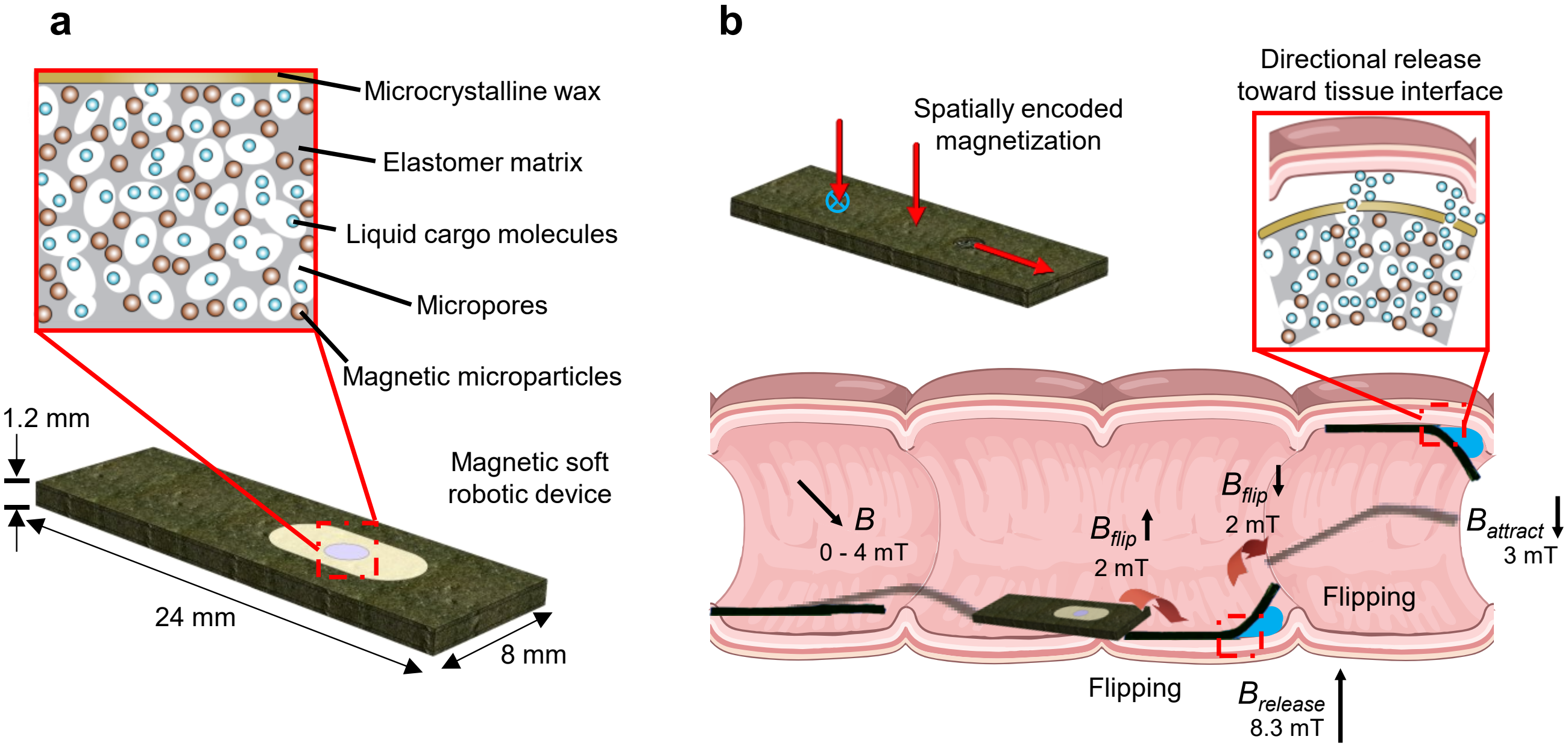

**Figure 1. Magnetic soft robot for directional release toward tissue interface with programmable locomotion and shape morphing. (a)** Magnetic soft robot embedded with interconnected microporous structure for liquid cargo storage, which is isolated by a sealing microcrystalline wax thin film. **(b)** Magnetization profile, and directional release toward tissue interface illustration with controlled locomotion and on-demand liquid cargo release under external dynamic magnetic fields.

## II. MAGNETIC SOFT MATERIAL FABRICATION

### A. *Concept*

The proposed magnetic soft robot for controlled directional release relies on a magnetization profile with one end and the middle sections oriented perpendicular to the longitudinal axis of the robot (Fig. 1b, top left). This magnetization configuration enables deterministic orientation control of the soft robot through reversal of the external magnetic field direction. For liquid cargo storage and isolation, an interconnected microporous elastomeric unit is employed to store aqueous cargos, while a thin microcrystalline wax film is laid out on the top surface of the soft robot to seal the release port for liquid cargo encapsulation (Fig. 1a). Triggered cargo release is achieved by mechanically rupturing the wax sealing layer under large bending deformation induced by higher external magnetic field. In contrast, lower magnetic fields generate smaller bending angles that enable crawling locomotion without breaking the sealing layer (Fig. 1b, bottom). This separation regime allows locomotion and cargo release to be independently controlled. The out of plane magnetization component within the robot body enables the controlled flipping of the magnetic soft robot by an opposing external magnetic field direction generated by a permanent magnet. This orientation control enables the alignment of the cargo release interface toward the desired direction, such as tissue interface in GI tract.

### B. *Fabrication of Microporous Magnetic Soft Robot*

The microporous magnetic soft robot was composed by assembling different parts together, including top sealing layer, central sealing frame, porous unit, and bottom sealing layer (Fig. 2a). The top, central, and bottom sealing material was made by Ecoflex-gel (33% w/w, part A and part B in 1:1 ratio, Smooth-on) mixed with silica-coated magnetic particles (average diameter of 5 µm, 67% w/w, Magnequench, MQP-15-7-20065). The porous unit also made by PDMS based elastomeric precursor, which was prepared by mixing 5.5 g fine (42% w/w) and 0.5 g ultra fine (4% w/w) sugar particles and 4 g silica-coated $Nd_2Fe_{14}B$ magnetic microparticles (31% w/w) into 3 g Ecoflex™ 00-30 (23% w/w, part A and part B in 1:1 ratio, Smooth-On). Sugar particles and magnetic particles were added first and mixed by using a Vortex (Analog Vortex Mixer, VMR) under 3200 RPM for 90 s before adding Ecoflex™ 00-30. Ultra fine sugar particles were produced by grinding coarse sugar particles for 2 minutes in an herb grinder. Sugar particles act as sacrificial porogen materials to generate porous internal structure [12], magnetic microparticles enabled magnetization and promote actuation properties under external magnetic field. Ecoflex 00-30 was added to the mixture and mixed for 3 minutes by using a milk frother.

For fabrication of microporous magnetic elastomer units, silicone precursor was poured into a template made of three sheets of double-sided tape on a glass slide which has been spin-coated with Polyvinyl Alcohol (PVA) (Sigma Aldrich) to prevent elastomer from adhering to substrate. Also, another spacer decorated with surface coated sugar particles on a top substrate, spacers with sugar decoration were silane coated by using Trichloro(1H,1H,2H,2H-perfluorooctyl)silane (Sigma Aldrich) in a vacuum chamber for 30 minutes to facilitate peeling off after curing. The sugar decorated top slide was pressed onto the precursor by applying same amount of loading force to ensure uniform and controlled thickness. Surface with sugar particles allows open air porous top surface, while the bottom surface is devoid of porous openings to enhance the bonding between the porous unit and bottom sealing Ecoflex-gel layer. The open air porous top surface is employed to enhance the drug loading efficiency and act as a support structure to the top sealing layer. The precursor is compressed uniformly between the two substrates by applying same load for different patches' fabrication to ensure consistent sample thickness and set aside to cure at room temperature for over 4 hours. The cured elastomers were then peeled off carefully and immersed in water to dissolve the sacrificial sugar particles for 48 hours at room temperature and then air dried to remove all remaining water molecules in the porous unit. The microporous magnetic elastomers were also dried under room temperature for 48 hours. The dried magnetic microporous elastomers were plasma treated (BD-20, Electro-technic products. Inc) for 3 minutes to enhance the liquid cargo wettability to facilitate the liquid cargo loading. Rectangle porous units with rounded corners and central sealing frame were excised by a metal punch for standardized fabrication. The total thickness of the magnetic soft robot was 1.2 mm. Both top and bottom sealing layers are 0.2 mm thick, and the porous unit and central sealing frame is 0.8 mm thick. The assembled robots were plasma treated for 2 minutes to increase the surface energy for enhancing adhesion between the wax seal layer and the robot surface.

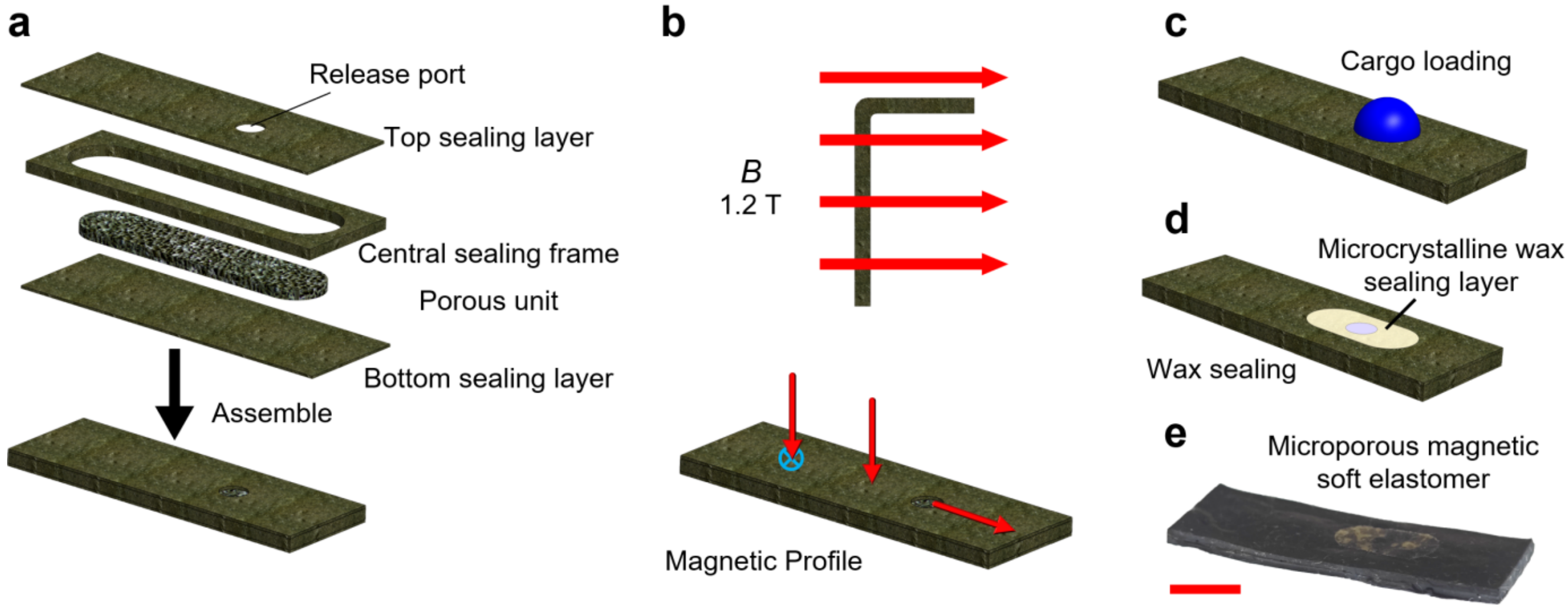

**Figure 2. Fabrication, programming, liquid cargo loading and wax sealing of microporous magnetic soft elastomers.** **(a)** Magnetic soft elastomer assembled by different PDMS based structures. **(b)** Spatial magnetic programming of magnetic soft elastomer. **(c)** Liquid cargo loading. **(d)** Microcrystalline wax film for sealing magnetic soft elastomers is achieved by directly laying and bonding to the top surface of the magnetic soft elastomers loaded with liquid cargo. **(e)** Final sample with loaded liquid cargo and wax sealing. Scale bar, 5 mm.

### *C. Magnetic Programming and Loading of Liquid Cargoes*

The magnetic soft robots were wrapped on a cube-shaped template and then exposed to a large, uniform magnetic field (1.2 T) in a pulse magnetizer (IM-10, ASC Scientific) to encode a spatially distributed magnetization profile as shown in Fig 2b. The distributed magnetization profile of magnetic soft robot after release from template is shown at the bottom panel of Fig 2 b. A water based blue food dye (Herbeklab) was used as a model drug and liquid cargo. Blue dye was loaded on the release port for 1 minute for absorption into the microporous internal structure of the magnetic soft elastomers (Fig. 2c).

### *D. Wax Film Preparation and Sealing of Liquid Cargo Loaded Magnetic Soft Elastomers*

A thin microcrystalline wax sealing layer with a 31 µm thickness was fabricated by molding microcrystalline wax (BW 42905, Blended Waxes, 73.9 °C- 79.4 °C drop melting point) into a channel template and then was heated to 80 °C on a hotplate to melt the wax. A PVA coated glass slide was used for uniform and intact debonding of the wax layer by dissolving the PVA layer in water over night. A 4 mm width and 7 mm length piece was then excised out of the thin wax sheet by punching. The wax thin film was directly laid on top of the magnetic soft robot to cover the release port for liquid cargo isolation (Fig. 2d). High resolution image of the final magnetic soft robot assembly is shown in Fig 2 e.

### *E. Actuation of Magnetically Programmed Soft Materials*

Actuation of the magnetic soft robot was achieved either using a permanent magnet or an electromagnetic coil. An N52 neodymium disc magnet (3" diameter x 1/8" thickness, KJ magnetics) was used to generate magnetic fields for flipping and locomotion characterization. The electromagnetic coil contains 240 windings of an 18-gauge enamel-coated copper wire around a square cross sectional ferrite core (25 x 25 mm$^2$) with a permeability of ~2500 (3C94, Ferroxcube). The electromagnetic coil setup used to induce the external magnetic field was directly connected by a power supply (HANMATEK HM310). The generated magnetic fields were measured with a magnetometer (PCE-MFM 3500, PCE Instruments).

### *F. Image and Video Recording*

High resolution images of the microstructure of fabricated microporous magnetic soft elastomers were captured using a stereomicroscope (Zeiss Stemi 305, 2x front optics objective) with an integrated eyepiece camera (Swiftcam SC2503). Magnetic actuation and aqueous dye release from magnetically programmed soft elastomers were recorded with a commercially available digital microscope (EdgePlus, Dino-Lite). Flipping characterization and ex-vivo walking experiment were recorded through a smartphone (Xiaomi 13 Ultra, Xiaomi). An endoscope was used for ex-vivo experiment recording in large intestine lumen. The porosity of microporous magnetic soft elastomers was quantified by ImageJ Fiji (National Institutes of Health). Photoshop (Adobe) was used for image processing to overlay different time-lapses for ex-vivo walking and top-substrate flipping experiments.

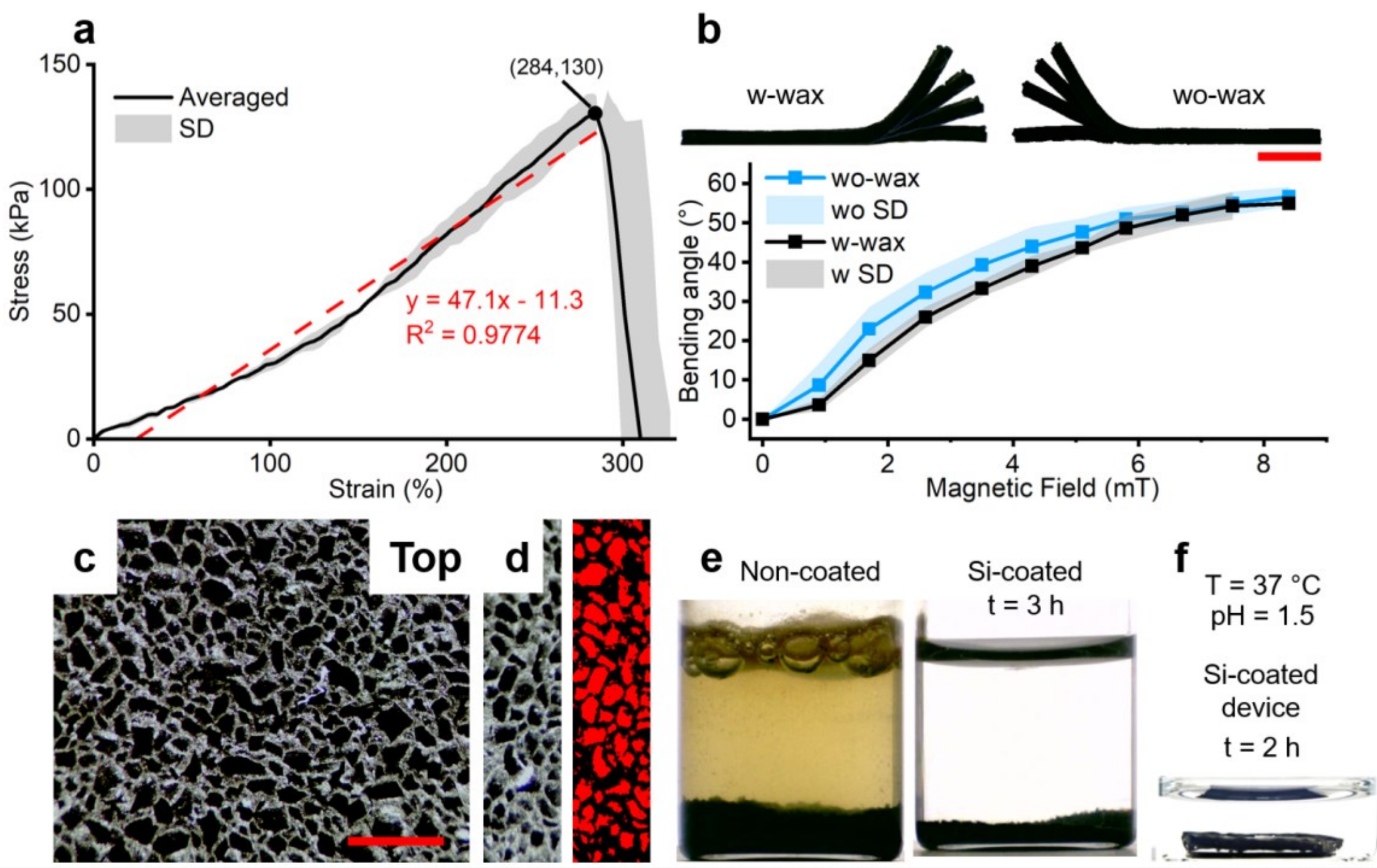

**Figure 3. Material characterization of microporous magnetic soft elastomers. (a)** Tensile test results for microporous elastomer samples. Averaged Young's Modulus is 47.1 kPa. **(b)** Magnetic soft elastomer bending angle comparison with and without wax sealing (Bending angles comparison under 0, 1.7 3.5, and 8.4 mT were illustrated). Wax sealing layer has mild inhibition for magnetic bending. Scale bar, 5mm. **(c)** Top surface of microporous magnetic elastomer before cargo loading. Scale bar, 500 µm. **(d)** Cross-section of microporous samples with an averaged porosity 49.3% was measured by Image J. **(e)** Low pH leaching test for non-coated and silica-coated (Si-coated) magnetic particles. With silica coating, magnetic particles didn't get any visible chemical corrosion in simulated gastric fluid environment for 3 hours. **(f)** Magnetic soft robot with liquid cargo loading and wax sealing did not show any visible cargo release under 37 °C for 2 h low pH leaching test.

### *G. Ex-vivo Experiment Swine sample*

A 2 feet length large intestine fresh tissue was collected from a 45 kg swine tissue (ID: BO569, Swine research and teaching center, University of Wisconsin-Madison). Large intestine tissue has a 3 cm diameter under inflated state. Tap water was used to flash the tissue for stool removing and washing.

## III. Experiments and Results

### *A. Morphological Characterization of Microporous Magnetic Soft Elastomers*

Mechanical properties of the microporous magnetic elastomers were characterized under tensile testing (KERN & SOHN GmbH) with 3 dog bone–shaped samples at a strain rate of 0.1316 $s^{-1}$. The results show a mean elastic modulus of ~ 47.1 kPa (Fig. 3a). Effect of wax sealing on deformability of magnetic soft robots was investigated by bending samples with and without wax sealing under increasing magnetic fields (Fig. 3b). While samples with wax sealing showed slightly lower bending angles compared to native samples, the difference decreased with increased magnetic fields and vanished above 6 mT (Fig. 3b). The reduced difference above 6 mT arises from the combined effects of magnetic field decay with increasing distance from the magnet during bending and the increased elastic resistance of the elastomer at larger deformation. The high-resolution microscopy images of the top porous surface and cross-section of the porous unit were shown in Fig. 3c and d. An average porosity of 49.3% ± 1.55% was measured from 3 different positions and 2 fabrication batches. We further investigated how the magnetic particles and the soft magnetic robot with cargo loading perform under simulated gastric fluid environment with 1.5 pH value at 37 °C. The leaching test showed the silica coating protected magnetic particles to avoid chemical corrosion under low-pH environment, and the wax sealing still maintained good isolation performance without any visible cargo leakage for 2 hours during the leaching experiment (Fig. 3e, f).

### *B. Magnetic Field-triggered Release of Liquid Cargo*

To characterize the sealing performance and releasing kinetics of the wax film, we experimentally measured the magnetic field required for breaking the sealing layer. An average of 8.37 ± 0.85 mT external magnetic field was needed to break the wax and release liquid cargo (Fig. 4a, b, Supplementary video). The current was incremented at 0.2 amp intervals from 0 amp (0 mT) until wax rupture. The same process with magnetic field actuation between 0 to 2 amps (0

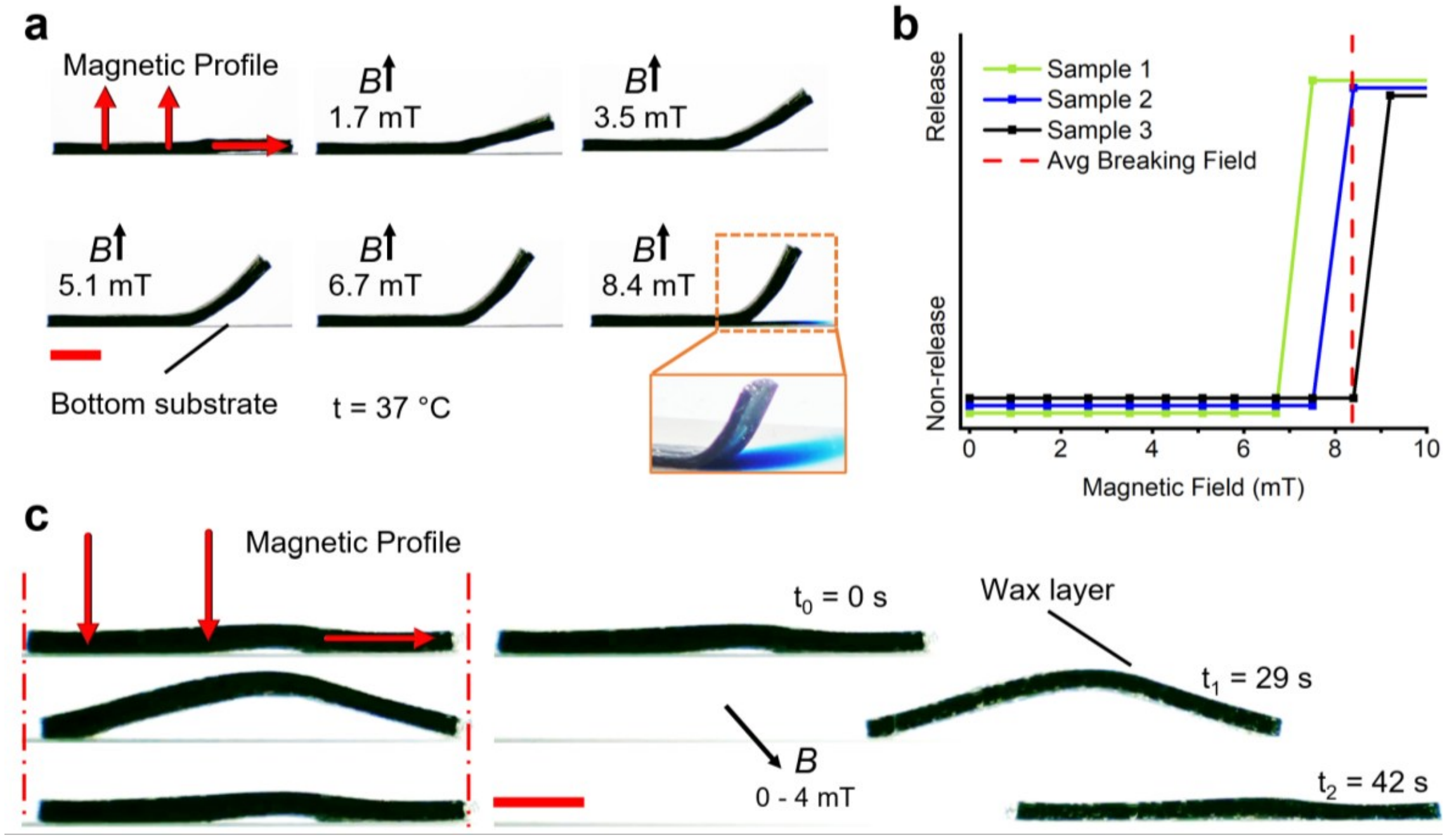

**Figure 4. Magnetic controlled releasing and locomotion characterization. (a)** Controlled liquid cargo release under external magnetic field actuation at 37 °C. **(b)** Qualification results of magnetic controlled releasing experiment with an averaged 8.37 mT breaking threshold was measured. **(c)** Crawling locomotion of soft magnetic robot under 0-4 mT dynamic external magnetic field. Red arrows indicated the magnetization profile. Scale bar 5 mm.

mT - 8.4 mT, respectively) was applied for the samples without wax sealing in the bending angle comparison experiment (Fig. 3b). At each increment, a time-lapse photo was captured from a recorded video. This experiment results also show there was not any visible liquid cargo leakage before reaching the threshold, which verified the wax isolation performance. The results indicate a large safety regime for locomotion operation under 5 mT.

### *C. Crawling Locomotion Characterization*

To further characterize the crawling locomotion of the magnetic microporous soft robot loaded with liquid cargo and sealed by wax thin film, we applied low frequency dynamic external magnetic field from 0 – 4 mT (Fig. 4c). There was no visible cargo leakage after cyclic crawling locomotion, which indicates the wax sealing performs well during transportation (Supplementary video).

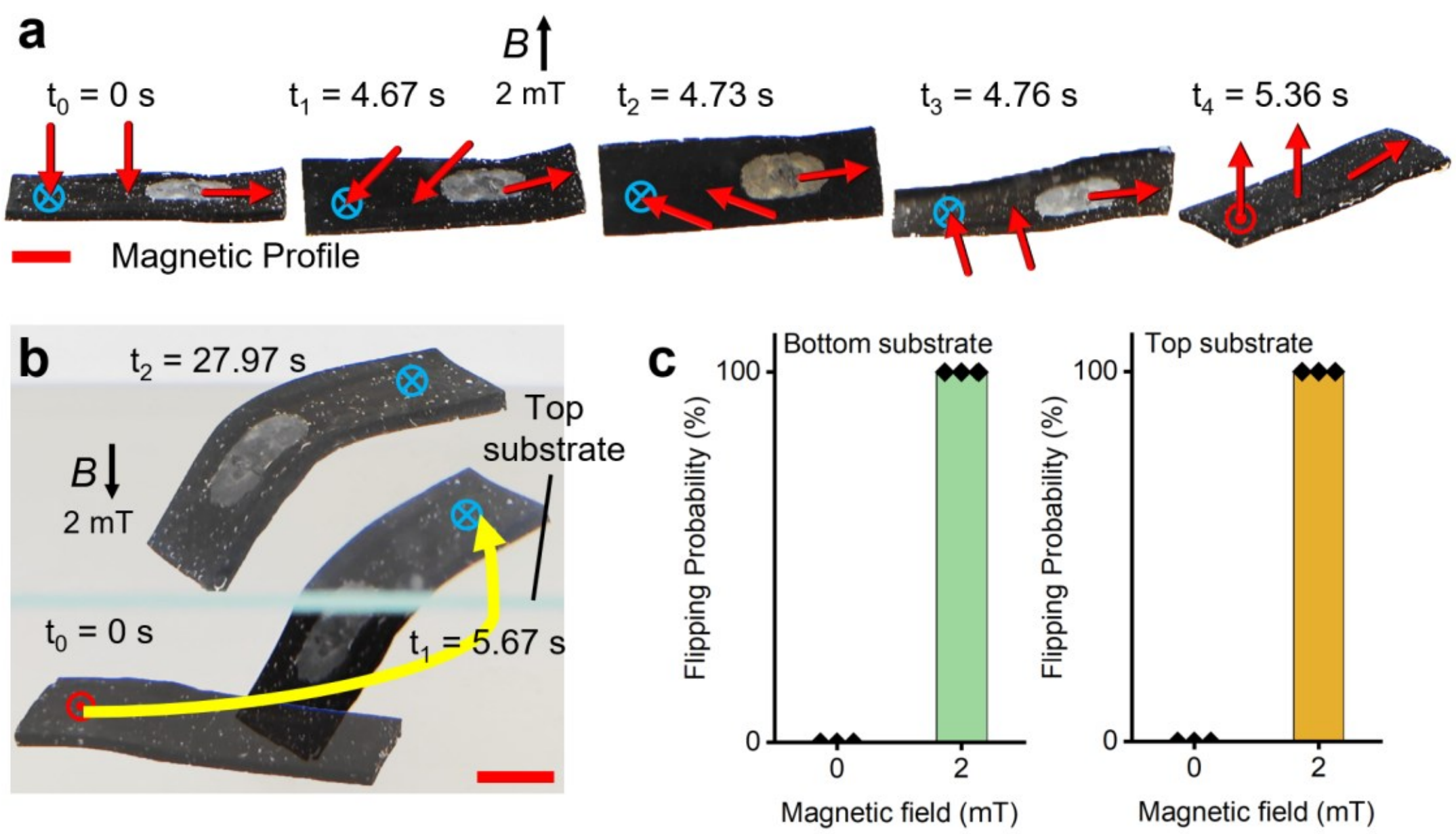

**Figure 5. Magnetic flipping characterization of magnetic soft elastomer for directional control. (a)** Flipping toward the bottom substrate illustration. Scale bar 5 mm. **(b)** Flipping toward the top substrate illustration. Scale bar 5 mm. **(c)** Flipping qualification results for both bottom and top substrate (3 samples, 10 times test each).

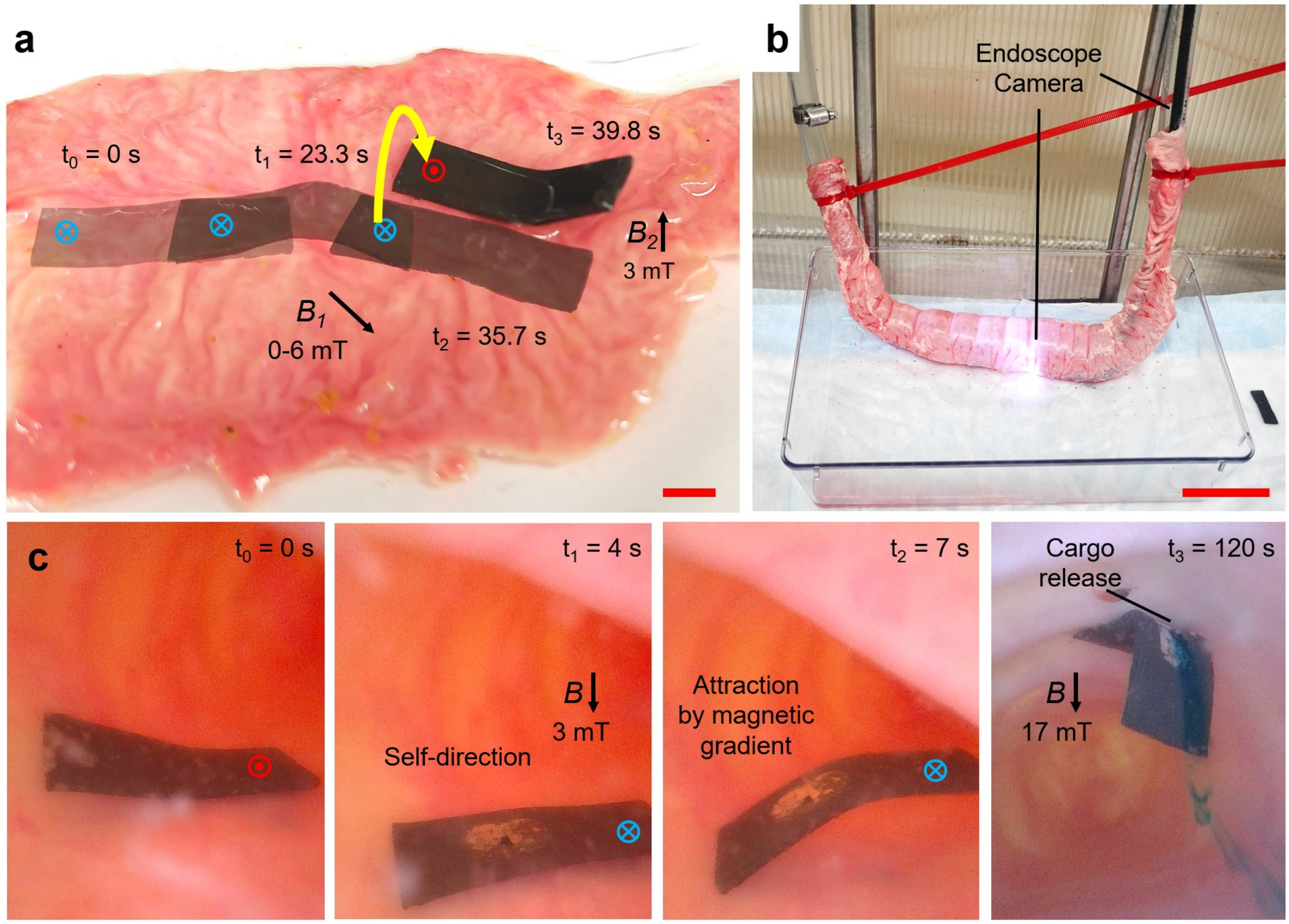

**Figure 6. *Ex-vivo* experiments for crawling, flipping, and directional release in swine large intestine tissue. (a)** Controlled crawling locomotion on a large intestine surface. Scale bar 5 mm. **(b)** Large intestine setup for flipping and releasing with endoscopy equipment. Scale bar 5 cm. **(c)** Endoscopy view of the large intestine for controlled directional release toward the tissue interface.

### D. *Controlled Flipping Characterization*

To characterize the self-correcting orientation function, the controlled flipping experiments were done by using 3 different samples with wax sealing for both top and bottom substrate flipping scenarios. For bottom substrate flipping scenario, a 2 mT magnetic field with an opposite magnetic orientation compared to the out of plane magnetization component of the magnetic soft robot was applied by approaching a N52 neodymium disc magnet. Different time-lapse images during one of the flipping experiments were captured from recorded video for illustration (Fig. 5a, Supplementary video). The out of plane magnetization component of the magnetic soft robot finally aligns with the external magnetic field after flipping.

To characterize the anti-gravity flipping performance of the robot and further understand its arbitrary orientation control capability, a top-substrate flipping experiment was conducted. Initially, the wax layer faced the bottom substrate but flipped with ease and faced towards the top substrate under an opposing magnetic field (2 mT) direction compared to the out of place magnetization profile of the magnetic soft robot. Different time-lapse images illustrated the flipping process (Fig. 5b). A quantification analysis was conducted by analyzing 3 different samples flipping 10 times each at both top and bottom substrate under 2 mT external magnetic field. The results show 100% successful flipping for all 3 samples out of 10 flipping cycles (Fig. 5c, Supplementary video). Both top and bottom substrate flipping require 2 mT external magnetic field, which is lower than the wax breaking threshold ~ 8 mT. The results show a safe operation regime for controlled orientation without rupturing the wax seal layer.

### E. *Ex-vivo Experiment for Controlled Locomotion and Controlled Directional Release.*

To further comprehensively demonstrate the controlled directional release capabilities with locomotion and on-demand target liquid cargo release of our magnetic microporous soft robot, *ex-vivo* experiments for crawling locomotion, flipping, and targeted cargo release were performed by using a large intestinal tissue from a 45 kg Swine. An opened large intestinal tissue was used for crawling locomotion and bottom substrate flipping demonstration (Fig. 6a, Supplementary video). The results indicate that the magnetic soft robot was able to crawl on a wrinkled tissue surface under a 0-6 mT low frequency dynamic magnetic field control, and flipping can be done by rotating external permanent magnet after controlled crawling to target position with a 3 mT magnitude (Fig. 6a). The *ex-vivo* experiment required a higher magnetic field for locomotion and flipping. This is attributed to the surface tension at the mucus-air interface, which generates additional interfacial resistance and slightly inhibited the bending deformation of the magnetic soft robot. The wrinkled surface of large intestine tissue also requires larger deformation to achieve locomotion.

In addition, we characterized the magnetic microporous soft robot's top substrate flipping and control directional release performance in a large intestine lumen environment. The large intestine lumen was filled with water, and an endoscope with LED light is used for video recording (Fig. 6b). The results indicate the magnetic soft robot's orientation can be controlled by flipping the orientation of the external permanent magnet, resulting in attraction and anchoring at the top tissue interface for continuous release under a 17 mT external magnetic field (Fig. 6c, Supplementary video).

## IV. CONCLUSION

In this paper, we present a new magnetic soft robot design with self-correcting capability for controlled and directional liquid cargo release. The introduction of the transverse magnetization component profile perpendicular to the longitudinal axis of the magnetic soft robot, enabled the controlled releasing orientation to the arbitrary interface of the intestinal tissue. The introduced design further preserves programmable locomotion and shape morphing capabilities. The flipping characterization and *ex-vivo* experiments show the robustness of the directional control. This work established the groundwork for effective target drug delivery in gastrointestinal tract application.